%% file: main.tex
\documentclass[11pt]{article}
\usepackage[preprint]{acl}
\usepackage{times}
\usepackage{latexsym}
\usepackage[T1]{fontenc}
\usepackage[utf8]{inputenc}
\usepackage{microtype}
\usepackage{inconsolata}
\usepackage{graphicx}
\usepackage{amsmath,amssymb}
\usepackage{booktabs}
\usepackage{tabularx}
\newcommand{\model}{Jev}
\newcommand{\va}{\mathrm{VA}}
\newcommand{\clip}{\operatorname{clip}_{[1,9]}}
\newcommand{\score}{\textsc{Score}}
\newcommand{\choice}{\textsc{Choice}}
\newcommand{\noul}{\textsc{Noul}}
\newcommand{\semevaltask}{SemEval-2026 \textbf{Task~III}}
\definecolor{oursrow}{HTML}{E8F0FB}
\definecolor{gain}{HTML}{1B7F3B}
\definecolor{weak}{HTML}{C2571A}

\makeatletter
\renewcommand{\@fnsymbol}[1]{\ifcase#1\or *\or **\or ***\fi}
\makeatother
\newcommand{\yes}{\scalebox{1.45}{$\bullet$}}
\newcommand{\no}{\textcolor{black!40}{\scalebox{1.45}{$\circ$}}}

\newcommand{\oracle}[1]{\textcolor{black!55}{+#1}}
\newcommand{\lowacc}[1]{\textcolor{weak}{#1}}
\newcommand{\shaderow}[1]{\rlap{\setlength{\fboxsep}{0pt}\colorbox{oursrow}{%
  \rule[-\dp\strutbox]{0pt}{\dimexpr\ht\strutbox+\dp\strutbox+1pt\relax}\hspace{#1}}}}
\title{Decide, Don't Generate:\protect\\Competitive Dimensional ABSA with Jev's Typed Decisions}
\author{Yiqun Zhang \quad Peidong Wang \quad Zihan Wang \quad Shi Feng\thanks{Corresponding author.} \\
  Northeastern University, China}
\begin{document}
\maketitle
\begin{abstract}
Aspect-based sentiment analysis (ABSA) has largely turned to text generation. We show that competitive dimensional ABSA does not need it. Using Jev, a frozen model that answers typed questions with rubric scores, label probabilities, and yes/no judgments, we decompose all three tasks of \semevaltask{} Track A into such decisions and align them with the annotation scheme through 488 coefficients fitted on CPU, with no text generation and no backbone tuning. On valence--arousal regression over ten corpora in six languages, the system reaches 1.0645 RMSE, the lowest aggregate error of any participating system. On triplet and quadruplet extraction, it reaches 52.09 and 44.06 continuous F1, above fine-tuned Llama-3.3-70B and GPT-OSS-120B baselines. Analyses and ablations show where the accuracy comes from: supervised calibration roughly halves the raw regression error, exact valence--arousal would add only 4.5 F1 to extraction, and the learned combination of span-boundary evidence, not any single signal, carries the extraction systems.\footnote{Code: \url{https://github.com/ZhangYiqun018/jev-dimabsa}}
\end{abstract}
\input{sections/introduction}
\input{sections/related-work}
\input{sections/method}
\input{sections/experiments}

\input{sections/conclusion}

\bibliography{references}
\appendix
\input{sections/appendix}

\end{document}

%% file: sections/introduction.tex
\section{Introduction}

\begin{figure}[t]
  \centering
  \includegraphics[width=\columnwidth]{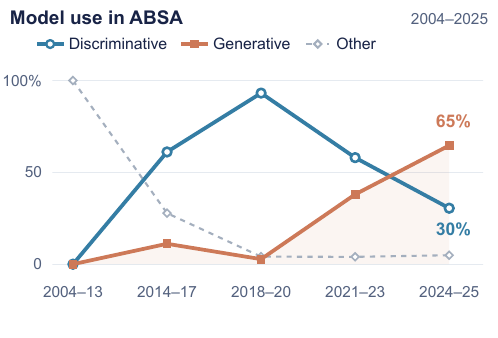}
  \caption{Share of ABSA methods using a text-generative model at any stage, by publication period (protocol in Appendix~\ref{app:literature-audit}).}
  \label{fig:absa-trend}
\end{figure}

Generating sentiment structures has become a prominent approach to aspect-based sentiment analysis (ABSA). Representative milestones include unified BART-based sequence prediction \citep{yan-etal-2021-unified}, T5-based paraphrase generation for aspect sentiment quadruples \citep{zhang-etal-2021-aspect-sentiment}, and multi-view prompting over output orders \citep{gou-etal-2023-mvp}. In a title-filtered audit of the literature (Figure~\ref{fig:absa-trend}), the share of methods that use a text-generative model rises from 38\% in 2021--2023 to about 65\% in 2024--2025. This count includes auxiliary uses such as augmentation and scoring, not only generation of the final tuples. The trend raises a question: \emph{does competitive sentiment analysis require generating text?}

\begin{figure*}[t]
  \centering
  \includegraphics[width=\textwidth]{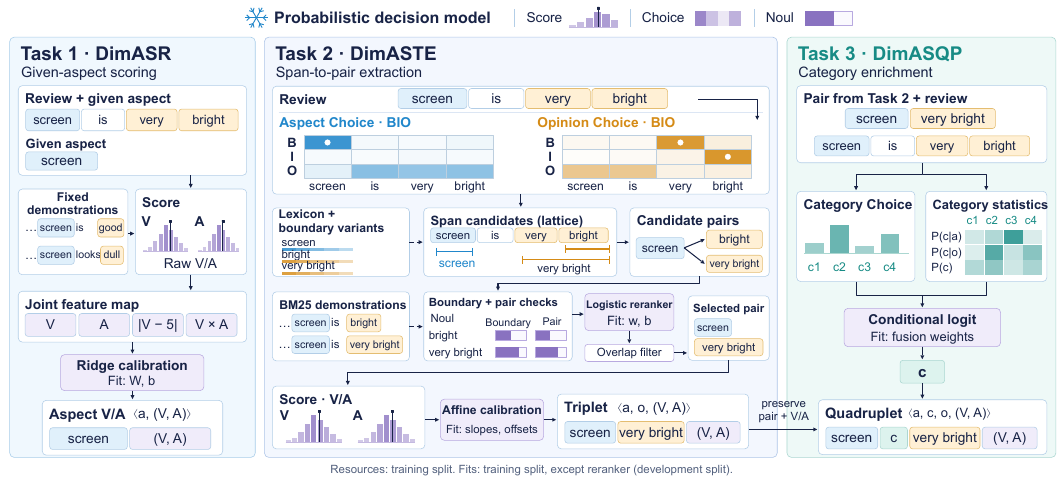}
  \caption{Pipeline for the three tasks. A frozen model (snowflake) answers every question with rubric scores, label probabilities, or yes/no judgments; ``Fit'' marks learned postprocessors, fitted on training data except the Task 2 reranker (development data). Task 1 calibrates the scores of a given aspect; Task 2 proposes, checks, and selects aspect--opinion pairs and scores their VA; Task 3 adds a category to each pair.}
  \label{fig:overview}
\end{figure*}

Before this shift, most ABSA systems were discriminative: an encoder supplied representations for classification or structured selection, as in BERT \citep{devlin-etal-2019-bert} and its sentence-pair formulation for ABSA \citep{sun-etal-2019-utilizing}. When the answer is a label, a span, or a number, direct prediction is a natural fit. Generation offers one output space for all of them, at the price of backbone adaptation, repeated decoding, and output validation.

TypeSafe's \emph{System One} framing gives a reason to revisit direct prediction. It presents \model{} as a model for fast, structured decisions: given a state and a typed question, it returns an answer with probabilities \citep{typesafe2026}. This resembles the role of BERT-style classifiers, with a question interface in place of task-specific output heads. We call the interface \emph{discriminative} without assuming anything about \model{}'s architecture or pretraining objective.

Our testbed is dimensional ABSA. It extends feature-level opinion mining \citep{hu2004mining} and polarity-based ABSA \citep{pontiki-etal-2014-semeval,pontiki-etal-2016-semeval} with continuous valence and arousal: how positive an evaluation is and how activated the expressed feeling is \citep{lee2026dimabsa}. \semevaltask{} (DimABSA, officially numbered Task~3) defines three increasingly structured tasks in its Track A \citep{yu-etal-2026-semeval}: given-aspect regression (DimASR), aspect--opinion triplet extraction (DimASTE), and categorized quadruplet prediction (DimASQP), which we call Tasks 1--3 (T1--T3 in tables). The latter two credit a tuple only when its spans match exactly, so they test structural prediction as well as numerical estimation.

Figure~\ref{fig:overview} shows our system, which decides rather than generates. Training-corpus statistics provide lexical candidates, boundary conventions, demonstrations, and category priors. \model{} provides rubric scores, token and category probabilities, and pair judgments. Ridge regression, logistic reranking, and a small fusion model align these decisions with the annotation scheme. \textbf{No component generates text, as an intermediate or a final answer, and no backbone weights are updated}; the system does use labeled data, through 488 coefficients fitted on CPU.

Our contributions are:
{\setlength{\topsep}{2pt}\setlength{\leftmargini}{1.1em}
\begin{itemize}\setlength{\itemsep}{1pt}\setlength{\parsep}{0pt}\setlength{\parskip}{0pt}
\item A decision-based pipeline for all three tasks that generates no text and tunes no backbone; its task adaptation is 488 coefficients fitted on CPU (Section~\ref{sec:method}).
\item The lowest ten-corpus T1 aggregate of any participating system (1.0645 RMSE), and T2/T3 scores (52.09/44.06 cF1) above fine-tuned Llama-3.3-70B and GPT-OSS-120B baselines, over six languages and four domains (Section~\ref{sec:main}).
\item Ablations and error analysis of where the accuracy comes from: calibration roughly halves the raw regression error, the learned combination of boundary evidence carries extraction (Section~\ref{sec:ablation}), and the remaining extraction error is structural: gold pairs are lost in proposing and in selecting spans, while adding categories costs us less than any leading system (Section~\ref{sec:structural}).
\end{itemize}}

%% file: sections/related-work.tex
\section{Related Work}
\subsection{ABSA formulations}
SemEval established aspect-level polarity prediction \citep{pontiki-etal-2014-semeval,pontiki-etal-2016-semeval}; DimABSA extends it to continuous valence--arousal ratings \citep{russell1980,lee2026dimabsa}, evaluated jointly with exact tuple structure through continuous F1 \citep{yu-etal-2026-semeval}. Discriminative approaches use BERT \citep{devlin-etal-2019-bert}, including sentence-pair classification \citep{sun-etal-2019-utilizing}. Generative approaches serialize sentiment structures through unified BART prediction \citep{yan-etal-2021-unified}, T5 paraphrases \citep{zhang-etal-2021-aspect-sentiment}, or multi-view output-order prompting \citep{gou-etal-2023-mvp}. We revisit direct prediction for all three dimensional tasks, constructing tuples through candidate selection rather than text decoding.

\subsection{DimABSA systems and calibration}
Published systems combine backbone adaptation (PAI, TeleAI, and PALI; \citealp{ruan-etal-2026-pai,zhou-etal-2026-teleai,chen-2026-pali}), retrieved demonstrations and ensembling (Takoyaki; \citealp{yamada-etal-2026-takoyaki}), or repeated structured generation (nchellwig; \citealp{hellwig-etal-2026-nchellwig}). TeamLasse separates generative extraction from encoder-based VA regression \citep{strothe-2026-teamlasse}; Table~\ref{tab:main} summarizes each system's adaptation. Our pipeline combines BM25-retrieved annotation examples \citep{robertson2009} with \model{}'s fixed \score, \choice, and \noul{} interfaces \citep{typesafe2026}. Unlike contextual calibration, which estimates answer bias from content-free inputs \citep{zhao2021calibrate}, our VA calibration fits benchmark labels: its gains rely on supervision as well as prompting.

%% file: sections/method.tex
\section{Method}
\label{sec:method}
\subsection{Tasks and decision interface}
Let $x$ be a review, $a$ an aspect, $o$ an opinion, and $c$ a category. Sentiment is a vector $y=(v,r)\in[1,9]^2$, where $r$ denotes arousal to distinguish it from aspect $a$. Task 1 predicts $y$ given $(x,a)$. Task 2 predicts a set of $(a,o,y)$ tuples given $x$. Task 3 predicts $(a,c,o,y)$ tuples. Explicit terms are substrings of the input; an implicit aspect uses the benchmark sentinel \texttt{NULL} where permitted.

All three tasks use the same model through three operations. \choice{} returns a distribution over a supplied finite label inventory. \noul{} returns a scalar judgment for a binary proposition. \score{} returns a distribution over ordered rubric levels and its expected zero-based index. For dimension $d$ with nine levels, we map that score to the benchmark scale as
\begin{equation}
  z_d = 1 + \sum_{k=0}^{8} k\,p_d(k\mid x,a,o),
  \label{eq:score}
\end{equation}
omitting $o$ for Task 1. Valence levels run from negative to positive evaluation and arousal levels from calm to activated feeling. Each question asks about the target aspect rather than the tone of the whole review. Figure~\ref{fig:overview} shows how the operations compose; Appendix~\ref{app:prompts} provides the prompt templates and complete scoring criteria.

\subsection{Task 1: given-aspect regression}
\paragraph{Fixed demonstrations.}
Each corpus uses nine fixed training examples: the earliest eligible record in each of nine equal-width valence bands, with empty bands filled in file order. The state contains the review and these labeled examples; two \score{} questions per aspect give raw valence and arousal. A BM25-retrieved alternative was tried on development data and not adopted (Section~\ref{sec:ablation}).

\paragraph{Joint calibration.}
A corpus-specific regression maps the raw outputs $z=(z_v,z_r)$ to benchmark labels. Define
\begin{equation}
  \phi(z) = [z_v,z_r,|z_v-5|,z_vz_r]^\top.
\end{equation}
After standardizing these features with training means $\mu$ and standard deviations $s$, the prediction is
\begin{equation}
  \hat y=\clip\!\left(W^\top\frac{\phi(z)-\mu}{s}+b\right).
  \label{eq:ridge}
\end{equation}
We fit $W\in\mathbb{R}^{4\times2}$ and $b\in\mathbb{R}^2$ by ridge regression with an unpenalized intercept. The extremity and interaction terms let predicted arousal depend on how extreme the valence is. Five-fold grouped cross-validation on a training sample selects the ridge penalty, and development data select the calibration family. The shrinkage variant in Table~\ref{tab:ablation} fits each dimension independently as $\bar y_d+\alpha_d(z_d-\bar z_d)$.

\subsection{Task 2: dimensional triplet extraction}
\paragraph{Token decisions and candidate lattice.}
Deterministic tokenization separates Chinese and Japanese characters, other words, and punctuation. For each role (aspect or opinion), \choice{} gives B/I/O probabilities per token. Candidates include the argmax BIO spans and alternative spans supported by the token marginals. For tokens $i$ through $j$, the lattice score is
\begin{equation}
\begin{split}
 L(i,j)={}&\big[p_i(B)+p_i(I)p_{i-1}(O)\big]\\
          &\cdot\prod_{k=i+1}^{j}p_k(I)\,[1-p_{j+1}(I)],
\end{split}
\end{equation}
with boundary values $p_0(O)=1$ and $p_{n+1}(I)=0$. This is a heuristic score, not a normalized span distribution. We keep spans of at most 12 tokens with $L(i,j)\geq0.2$, add variants suggested by training boundary statistics, and include literal matches from the training lexicon. Each non-overlapping aspect--opinion combination receives a \noul{} pair judgment.

\paragraph{Boundary checks and extensions.}
A pair judgment says whether a relation is plausible, but the metric also requires the dataset's exact boundaries. Further \noul{} questions therefore ask whether a candidate is exactly one annotated phrase and whether a pair follows the dataset's relation convention. Their context includes four same-corpus training reviews retrieved by BM25. Each check is repeated with character-bigram, character-trigram, and word retrieval, using Jieba for Chinese words. Opinion candidates are also extended by up to three tokens to the left and two to the right, with no internal punctuation and at most 12 tokens; an extension enters the pair pool when its span check is at least 0.5.

\paragraph{Learned pair selection.}
For each candidate whose lattice-stage pair judgment is at least 0.3, a feature vector $f(x,a,o)$ combines model judgments, BIO support, boundary checks, training counts, edge statistics, length, distance, competing variants, extension membership, the mean and minimum check logits across retrieval views, and each check relative to the strongest overlapping rival. A logistic reranker predicts
\begin{equation}
  q(a,o\mid x)=\sigma\big(w_g^\top f(x,a,o)+b_g\big),
\end{equation}
where $g$ is a language group: English, Chinese, Japanese, or a shared Russian/Tatar/Ukrainian group. Each group model is fitted on development labels with an L2 penalty of 5, and features also encode corpus identity. Greedy decoding takes candidates in descending score, stops below 0.25, and drops a pair only when both its aspect and its opinion overlap an already selected pair, so one aspect can pair with several opinions and vice versa.

\paragraph{Pair-conditioned VA.}
For each selected pair, two \score{} questions estimate VA conditioned on both $a$ and $o$. Instead of Task 1's ridge model, an affine map per corpus $m$ and dimension, fitted on about 1,000 training gold pairs, gives
\begin{equation}
  \hat y_d=\clip(\beta_{m,d}z_d+\gamma_{m,d}).
\end{equation}

\subsection{Task 3: category enrichment}
Task 3 takes the predicted Task 2 pairs and their VA values. For each pair, \choice{} scores the corpus's category inventory. Each option names an entity and describes its attribute; the state includes four retrieved annotated reviews and a glossary of the three most frequent training aspects of each category.

We also estimate category priors and lexical distributions from training counts. With add-one prior $\pi_c$, the aspect lookup is
\begin{equation}
  P(c\mid a)=\frac{n(a,c)+\pi_c}{n(a)+1},
\end{equation}
and the opinion lookup is analogous. Surface forms are lowercased, and \texttt{NULL} has no lexical counts. A six-dimensional vector $h_c$ contains the log model probability, the log aspect lookup and its seen-in-training interaction, the log prior, and the log opinion lookup and its seen interaction. Model probabilities are floored at $10^{-3}$ before taking logs. A conditional-logit model gives
\begin{equation}
  P(c\mid x,a,o)=\frac{\exp(u_g^\top h_c)}{\sum_{c'}\exp(u_g^\top h_{c'})}.
\end{equation}
The six weights $u_g$ are fitted on about 1,000 training gold pairs per corpus, with L2 penalty 1 and no intercept. During fitting, counts exclude all annotations sharing the sampled review's normalized text, and retrieval excludes that text. The highest-scoring category is attached to the pair without changing its spans or VA.

%% file: sections/experiments.tex
\section{Experiments}
\label{sec:experiments}
\input{tables/main-results}

\input{tables/per-corpus}

\subsection{Setup}
\label{sec:protocol}
\paragraph{Data.}
We use the public Track A splits of \semevaltask{} \citep{yu-etal-2026-semeval}. Task 1 covers ten corpora in English, Chinese, Japanese, Russian, Tatar, and Ukrainian across restaurants, laptops, hotels, and finance; Tasks 2 and 3 cover the eight non-finance corpora. The Task 1 test set has 9,658 reviews and 16,186 aspect annotations; the Task 2 and Task 3 test sets share 6,690 reviews with 14,262 triplets and 14,263 quadruplets (Appendix~\ref{app:data}).

\paragraph{Metrics.}
We use the unchanged official scorer with Task 1 normalization disabled. Its joint error is
\begin{equation}
 \operatorname{RMSE}_{\va}=\sqrt{\frac{1}{N}\sum_{i=1}^N\|\hat y_i-y_i\|_2^2},
\end{equation}
pooled over all annotations of the ten corpora (\emph{micro}). For Tasks 2 and 3, each exact structural match contributes
\begin{equation}
 t_i=\max\left(0,1-\frac{\|\hat y_i-y_i\|_2}{\sqrt{128}}\right),
\end{equation}
where a match requires aspect and opinion, plus category for Task 3. Continuous precision and recall divide the summed contributions by the predicted and gold tuple counts; continuous F1 (cF1) is their harmonic mean, reported $\times100$ and averaged over the eight corpora.

\paragraph{Protocol.}
We use \texttt{jev-1.13.0} with the official data and scorer (Appendix~\ref{app:repro}). Every fitted component uses training labels except the Task 2 reranker, which is trained on development labels and scored out of fold. Development data also select all design choices and thresholds, and test labels are used only for evaluation; Appendix~\ref{app:repro} gives sampling and selection details. Retrieved examples exclude training texts that occur in development or test.

\subsection{Competitive without generation or tuning}
\label{sec:main}
Table~\ref{tab:main} compares our system with six participant systems and three model baselines evaluated by \citet{lee2026dimabsa}. It is the only system that neither tunes a backbone nor generates text in any task: its whole task adaptation is 488 coefficients fitted on CPU (100 in the Task 1 ridge models, 332 in the Task 2 rerankers, 32 in the pair-level VA maps, and 24 in category fusion). Because the competition ranks each corpus separately, we reconstruct participant aggregates from the published per-corpus scores. The systems also differ in backbone and supervision, so the comparison places our results in context rather than isolating the effect of generation.

\paragraph{Regression.}
Our system obtains 1.0645 micro RMSE, the lowest aggregate among the 14 teams that report all ten corpora and 0.0018 below PAI's 1.0663 \citep{ruan-etal-2026-pai}. Without participant predictions this margin cannot be tested for significance. Per corpus (Table~\ref{tab:corpora}), our system beats the best published score on English laptop and Tatar restaurant and trails it on the other eight.

\paragraph{Extraction.}
Task 2 reaches 52.09 cF1, between the sixth and seventh of the 12 complete-coverage participants, and Task 3 reaches 44.06, between the fifth and sixth of 9. \textbf{Both exceed the fine-tuned Llama-3.3-70B and GPT-OSS-120B baselines.} The two other systems without backbone tuning both generate text: our system trails Takoyaki's retrieval-and-rules pipeline by about 4 points on each task and exceeds one-shot Kimi K2 Thinking by 13.5 and 17.1 points.

\subsection{What each component contributes}
\label{sec:ablation}
\input{tables/ablation}
Table~\ref{tab:ablation} removes one component at a time from each final system. Each variant refits its learned postprocessor on the same data as the final system (training data for Tasks 1 and 3, development data for the Task 2 reranker); the ablations are post hoc and informed no design choice. For Task 2 we report exact-match pair F1, i.e., cF1 with gold VA, which isolates the structural decisions being ablated.

\paragraph{Task 1: calibration matters most.}
Raw scores reach only 2.0731 RMSE even with nine demonstrations: \textbf{nine rubric levels do not by themselves put the model's scores on the gold scale, but a few coefficients per corpus do.} Calibration removes about half of the error. Its joint terms, which let predicted arousal depend on how extreme the valence is, are worth 0.0558 over independent shrinkage, and the demonstrations add 0.0370 once scores are calibrated. On development data the joint terms improve RMSE by 0.0506, with a paired bootstrap 95\% interval of $[-0.0626,-0.0395]$, while retrieving examples with BM25 instead of fixing them does not help (0.8613 against 0.8572).

\paragraph{Task 2: boundary decisions matter most.}
Replacing the reranker, and all the evidence it combines, by the lattice pair judgment alone (with a threshold chosen on development data) loses 15.31 points: \textbf{no single signal decides boundaries well; their learned combination does.} Restricting candidates to the argmax BIO spans loses 6.09, the value of the lattice, and removing the example-conditioned checks loses 2.41. The extra retrieval views, the lexicon pair answers, and the opinion extensions each add less than a point.

\paragraph{Task 3: the model's category decision carries the signal.}
Without the model's category probabilities, training lookups and the prior reach only 38.03 cF1 ($-$6.03). Without the lookups, the model decision and prior alone match the full fusion (44.08 against 44.06).

\subsection{Remaining extraction error is structural}
\label{sec:structural}
\input{tables/errors}
\input{tables/retention}
\paragraph{Spans, not sentiment values.}
Gold VA on our extracted pairs would add only 4.46 cF1 to Task 2 (Table~\ref{tab:corpora}), so most of the error lies in which spans are extracted. Table~\ref{tab:errors} locates it. Averaged over corpora, 21.7\% of the gold pairs are never proposed as candidates and 26.8\% are proposed but not selected, and 42.7\% of the wrongly selected pairs are boundary near-misses that overlap a gold pair on both roles. The two losses split by language: the Russian, Tatar, and Ukrainian corpora lose more than a quarter of their gold pairs before selection, whereas the Chinese and Japanese corpora lose over a third among proposed candidates. Chinese laptop, our weakest corpus (39.43 against PALI's 53.08; \citealp{chen-2026-pali}), suffers from both, and two thirds of its false positives are near-misses. \textbf{The remaining gap lies in candidate coverage and boundary selection.}

\paragraph{Categories cost no more than for the leading systems.}
Adding a category lowers our macro cF1 from 52.09 to 44.06. This loss of 8.03 points is the smallest among the systems that match or exceed us on both tasks (Table~\ref{tab:retention}), so our 5.14-point gap to PALI on Task 3 is inherited from the Task 2 pairs. Category accuracy on matched pairs exceeds 91\% in every restaurant corpus but is 60.2\% in English laptop and 75.3\% in Japanese hotel, whose inventories have 113--121 and 44 labels.

%% file: tables/main-results.tex
\begin{table*}[t]
\centering\small
\setlength{\tabcolsep}{2.7pt}
\begin{tabular*}{\textwidth}{@{\extracolsep{\fill}}llccccc@{}}
\toprule
System & Task adaptation & \begin{tabular}[c]{@{}c@{}}Tunes\\backbone\end{tabular} & \begin{tabular}[c]{@{}c@{}}Generates\\text\end{tabular} & \begin{tabular}[c]{@{}c@{}}T1\\RMSE $\downarrow$\end{tabular} & \begin{tabular}[c]{@{}c@{}}T2\\cF1 $\uparrow$\end{tabular} & \begin{tabular}[c]{@{}c@{}}T3\\cF1 $\uparrow$\end{tabular} \\
\midrule
\multicolumn{7}{@{}l}{\emph{SemEval-2026 participants}} \\
PAI~\citep{ruan-etal-2026-pai} & LoRA + VA alignment & \yes & \yes & 1.0663 & \textbf{57.73} & -- \\
TeleAI~\citep{zhou-etal-2026-teleai} & LoRA + regression head & \yes & \yes\rlap{\,\footnotesize T2/3} & 1.0737 & 55.66 & 31.26 \\
PALI~\citep{chen-2026-pali} & LoRA adapters & \yes & \yes & 1.1340 & 57.50 & \textbf{49.20} \\
Takoyaki~\citep{yamada-etal-2026-takoyaki} & Retrieval + rules & \no & \yes & -- & 56.20 & 48.03 \\
nchellwig~\citep{hellwig-etal-2026-nchellwig} & LoRA & \yes & \yes & -- & 56.55 & 47.19 \\
TeamLasse~\citep{strothe-2026-teamlasse} & LoRA + encoder regressor & \yes & \yes\rlap{\,\footnotesize T2/3} & -- & 53.43 & 44.33 \\
\midrule
\multicolumn{7}{@{}l}{\emph{Model baselines of \citet{lee2026dimabsa}}} \\
Llama-3.3-70B~\citep{meta2024llama33} & 4-bit QLoRA & \yes & \yes & 2.5683 & 46.40 & 38.62 \\
GPT-OSS-120B~\citep{openai2025gptoss} & 4-bit QLoRA & \yes & \yes & 1.2362 & 45.71 & 37.27 \\
Kimi K2 Thinking~\citep{moonshot2025kimithinking} & One-shot prompting & \no & \yes & 1.8873 & 38.59 & 26.95 \\
\midrule
\shaderow{\textwidth}\textbf{Ours} & \textbf{488 coefficients on CPU} & \no & \no & \textbf{1.0645} & 52.09 & 44.06 \\
\bottomrule
\end{tabular*}
\caption{Test results and task adaptation. T1: micro RMSE over ten corpora; T2/T3: macro cF1 over eight corpora. \yes{} yes, \no{} no. Participant aggregates are computed from the per-corpus scores of \citet[Tables 6--8]{yu-etal-2026-semeval}; --: not every corpus reported. Best score per column in bold.}
\label{tab:main}
\end{table*}

%% file: tables/per-corpus.tex
\begin{table*}[t]
\centering\small
\setlength{\tabcolsep}{3pt}
\begin{tabular*}{\textwidth}{@{\extracolsep{\fill}}lcccccccc@{}}
\toprule
& \multicolumn{2}{c}{Task 1: RMSE $\downarrow$} & \multicolumn{3}{c}{Task 2: cF1 $\uparrow$} & \multicolumn{3}{c}{Task 3: cF1 $\uparrow$} \\
\cmidrule(lr){2-3}\cmidrule(lr){4-6}\cmidrule(l){7-9}
Corpus & Ours & Best & Ours & Best & Exact VA & Ours & Best & Cat.\ acc. \\
\midrule
English restaurant & 1.2163 & 1.1035\textsuperscript{a} & 68.21 & 70.21\textsuperscript{f} & \oracle{5.31} & 63.48 & 65.14\textsuperscript{f} & 93.0 \\
English laptop & \textbf{1.2086} & 1.2408\textsuperscript{a} & 62.25 & 63.66\textsuperscript{f} & \oracle{5.54} & 37.38 & 42.27\textsuperscript{f} & \lowacc{60.2} \\
Japanese hotel & 0.6454 & 0.5561\textsuperscript{b} & 50.03 & 58.37\textsuperscript{b} & \oracle{2.60} & 37.59 & 42.52\textsuperscript{g} & \lowacc{75.3} \\
Japanese finance & 0.7296 & 0.6581\textsuperscript{b} & -- & -- & -- & -- & -- & -- \\
Russian restaurant & 1.3290 & 1.2190\textsuperscript{c} & 51.26 & 57.93\textsuperscript{c} & \oracle{5.64} & 46.80 & 55.99\textsuperscript{c} & 91.3 \\
Tatar restaurant & \textbf{1.4604} & 1.5294\textsuperscript{c} & 45.09 & 51.19\textsuperscript{h} & \oracle{5.61} & 42.03 & 47.36\textsuperscript{f} & 93.1 \\
Ukrainian restaurant & 1.3464 & 1.1888\textsuperscript{c} & 50.24 & 57.87\textsuperscript{c} & \oracle{5.65} & 46.84 & 54.37\textsuperscript{c} & 93.2 \\
Chinese restaurant & 0.9591 & 0.9256\textsuperscript{d} & 50.21 & 56.38\textsuperscript{c} & \oracle{3.29} & 46.51 & 55.21\textsuperscript{i} & 92.6 \\
Chinese laptop & 0.7611 & 0.6103\textsuperscript{b} & 39.43 & 53.08\textsuperscript{g} & \oracle{2.07} & 31.88 & 48.24\textsuperscript{i} & 80.8 \\
Chinese finance & 0.5823 & 0.4841\textsuperscript{e} & -- & -- & -- & -- & -- & -- \\
\midrule
\shaderow{\textwidth}\textbf{Aggregate} & \textbf{1.0645} & 1.0663\textsuperscript{c} & \textbf{52.09} & 57.73\textsuperscript{c} & \oracle{4.46} & \textbf{44.06} & 49.20\textsuperscript{g} & 85.0 \\
\bottomrule
\end{tabular*}
\caption{Per-corpus test results. Best: best published score for the corpus, from \citet{yu-etal-2026-semeval}; for the aggregate, the best complete-coverage aggregate of Table~\ref{tab:main}. Superscripts: (a) LogSigma \citep{hikal-etal-2026-logsigma}, (b) TeleAI \citep{zhou-etal-2026-teleai}, (c) PAI \citep{ruan-etal-2026-pai}, (d) ICT-NLP \citep{huang-etal-2026-ict}, (e) HUS@NLP-VNU \citep{cao-etal-2026-hus}, (f) Takoyaki \citep{yamada-etal-2026-takoyaki}, (g) PALI \citep{chen-2026-pali}, (h) nchellwig \citep{hellwig-etal-2026-nchellwig}, (i) NYCU Speech Lab. Bold: better than Best. Exact VA: gain if our extracted pairs had gold VA. Cat.\ acc.: category accuracy (\%) on predicted pairs that match a gold pair, highlighted below 80. The finance corpora have Task 1 data only.}
\label{tab:corpora}
\end{table*}

%% file: tables/ablation.tex
\begin{table}[t]
\centering\small
\begin{tabular*}{\columnwidth}{@{\extracolsep{\fill}}lcc@{}}
\toprule
Variant (test) & Score & $\Delta$ \\
\midrule
\multicolumn{3}{@{}l}{\emph{Task 1, micro RMSE $\downarrow$}} \\
\shaderow{\columnwidth}Full system & \textbf{1.0645} &  \\
\quad $-$ joint terms (shrinkage) & 1.1203 & \textcolor{weak}{+0.0558} \\
\quad $-$ demonstrations (zero-shot) & 1.1015 & \textcolor{weak}{+0.0370} \\
\quad $-$ calibration (raw scores) & 2.0731 & \textcolor{weak}{+1.0086} \\
\midrule
\multicolumn{3}{@{}l}{\emph{Task 2, exact-match pair F1 $\uparrow$}} \\
\shaderow{\columnwidth}Full system & \textbf{56.55} &  \\
\quad $-$ opinion extensions & 56.47 & \textcolor{black!55}{$-$0.08} \\
\quad $-$ extra retrieval views & 56.03 & \textcolor{black!55}{$-$0.52} \\
\quad $-$ lexicon pair answers & 56.22 & \textcolor{black!55}{$-$0.33} \\
\quad $-$ example-conditioned checks & 54.14 & \textcolor{weak}{$-$2.41} \\
\quad $-$ lattice (argmax spans only) & 50.46 & \textcolor{weak}{$-$6.09} \\
\quad $-$ reranker (pair judgment only) & 41.24 & \textcolor{weak}{$-$15.31} \\
\midrule
\multicolumn{3}{@{}l}{\emph{Task 3, cF1 $\uparrow$}} \\
\shaderow{\columnwidth}Full system & \textbf{44.06} &  \\
\quad $-$ training lookups & 44.08 & \textcolor{black!55}{+0.02} \\
\quad $-$ model category decision & 38.03 & \textcolor{weak}{$-$6.03} \\
\bottomrule
\end{tabular*}
\caption{Component ablations on test (T1 micro over ten corpora, T2/T3 macro over eight). Each variant removes one component and refits the learned postprocessor on the data the final system uses. Removing the T2 checks also removes the extensions they admit. Losses of at least 0.01 RMSE or one F1 point are highlighted.}
\label{tab:ablation}
\end{table}

%% file: tables/errors.tex
\begin{table}[t]
\centering\small
\setlength{\tabcolsep}{3pt}
\begin{tabular*}{\columnwidth}{@{\extracolsep{\fill}}lcccc@{}}
\toprule
Corpus & \begin{tabular}[c]{@{}c@{}}Not\\proposed\end{tabular} & \begin{tabular}[c]{@{}c@{}}Not\\selected\end{tabular} & Found & \begin{tabular}[c]{@{}c@{}}Near-miss\\FP\end{tabular} \\
\midrule
Eng.\ rest. & \textbf{15.5} & 14.4 & 70.0 & 46.6 \\
Eng.\ laptop & 17.1 & \textbf{20.7} & 62.2 & 46.7 \\
Jpn.\ hotel & 17.2 & \textbf{34.5} & 48.3 & 36.3 \\
Rus.\ rest. & \textbf{26.0} & 21.3 & 52.7 & 32.5 \\
Tat.\ rest. & 27.2 & \textbf{28.5} & 44.4 & 30.4 \\
Ukr.\ rest. & \textbf{27.9} & 20.3 & 51.8 & 34.0 \\
Zho.\ rest. & 14.0 & \textbf{37.2} & 48.8 & 49.0 \\
Zho.\ laptop & 28.6 & \textbf{37.3} & 34.1 & 66.3 \\
\midrule
\shaderow{\columnwidth}\textbf{Macro} & 21.7 & 26.8 & 51.5 & 42.7 \\
\bottomrule
\end{tabular*}
\caption{Where Task 2 loses gold pairs on test (\%): never proposed as a candidate, proposed but not selected, or found; the larger loss per corpus is in bold. Near-miss FP: share of wrongly selected pairs whose aspect and opinion both overlap one gold pair.}
\label{tab:errors}
\end{table}

%% file: tables/retention.tex
\begin{table}[t]
\centering\small
\setlength{\tabcolsep}{3pt}
\begin{tabular*}{\columnwidth}{@{\extracolsep{\fill}}lccc@{}}
\toprule
System & T2 & T3 & Loss \\
\midrule
PALI~\citep{chen-2026-pali} & 57.50 & 49.20 & 8.30 \\
Takoyaki~\citep{yamada-etal-2026-takoyaki} & 56.20 & 48.03 & 8.17 \\
nchellwig~\citep{hellwig-etal-2026-nchellwig} & 56.55 & 47.19 & 9.36 \\
TeamLasse~\citep{strothe-2026-teamlasse} & 53.43 & 44.33 & 9.10 \\
\shaderow{\columnwidth}\textbf{Ours} & 52.09 & 44.06 & \textbf{8.03} \\
\bottomrule
\end{tabular*}
\caption{Macro cF1 on Tasks 2 and 3, and the loss when a category is added to the extracted pairs, for the systems that report every corpus of both tasks and score at least as high as ours on both. Smallest loss in bold.}
\label{tab:retention}
\end{table}

%% file: sections/conclusion.tex
\section{Conclusion}
Deciding instead of generating is enough for competitive dimensional ABSA. With \model{}'s typed decisions, corpus statistics, and 488 coefficients fitted on CPU, our system obtains the lowest ten-corpus Task 1 aggregate of any participating system and outperforms fine-tuned Llama-3.3-70B and GPT-OSS-120B baselines on extraction, without text generation or backbone tuning. \textbf{What makes it work is alignment with the annotation scheme}: a few coefficients per corpus put the model's scores on the gold scale, the remaining extraction error lies in proposing and selecting span boundaries rather than in sentiment values or categories, and ablations show that the learned combination of boundary evidence carries extraction. Closing that gap, and comparing the latency and cost of decision composition with generative systems under matched conditions, are the natural next steps.

%% file: sections/appendix.tex
\section{Dataset Details}
\label{app:data}
\input{tables/dataset}
Table~\ref{tab:data} reports counts of the official data before task-specific training filters. Training files for the non-finance corpora contain quadruplet annotations reused across tasks. Their counts therefore differ from the number of distinct Task 1 aspect targets. In English laptop test, Task 3 has 1,975 quadruplets and Task 2 has 1,974 triplets, so each task is evaluated against its own gold file. Task 3 reuses \emph{predicted} Task 2 pairs, which does not require the gold inventories to agree.

\section{Implementation and Reproducibility}
\label{app:repro}
We use \texttt{jev-1.13.0} and the official DimABSA data and scorer at commit
\begin{quote}\small\ttfamily\raggedright
bdc93be1224106ae7d3eb9\\
5739c02a76ed4ae8a1
\end{quote}
of the task repository. External comparisons use the published, rounded scores.

\paragraph{Supervision and selection.}
Task 1 demonstrations, the Task 1 ridge models, the Task 2 lexicon, boundary statistics and affine VA maps, and the Task 3 lookups and fusion weights are fitted on training labels. The Task 2 logistic reranker is trained on development labels; five-fold out-of-fold predictions by record give its development scores. Because features and thresholds were also chosen on development data, these scores are not nested estimates of the full selection procedure, and the folds do not group parallel Russian/Tatar/Ukrainian translations. The final Task 2 revision (opinion extensions, three retrieval views, and rival features) gained 1.50 cF1 out of fold but 0.41 on test; it changed several components at once.

Task 1 calibration uses 256 training text groups per corpus, 2,563 records and 4,664 VA annotations in total. Repeated texts stay in one group, and parallel translations share groups and folds. The sample excludes the fixed demonstrations and any text whose ID or normalized form occurs in development or test. The seed is 20260923, and ridge penalties are selected from $\{0.1,1,3,10,30,100,300\}$. A new calibration replaces the previous one only if it improves development RMSE by at least 0.02 and a paired cluster bootstrap (2,000 draws) puts the 95\% interval of the change below zero.

\paragraph{Overlap handling.}
The Task 1 audit finds 23 train--test text overlaps in Japanese hotel; the Task 2 subset has two normalized-text overlaps with train. Retrieved examples exclude training texts that occur in development or test, and Task 3 applies the same exclusion to its statistics and glossary. Task 2's lexicon and boundary counts use the full training split, so overlap removal is incomplete for that task.

\paragraph{Task 2 candidate details.}
Token BIO questions are batched at 48 questions per request, lattice pair checks at 32 pairs, and pair-conditioned VA at 16 pairs. The BIO state contains two synthetic examples. Training edge-affix variants use an inclusion/exclusion proportion of at least 0.9 with support of at least 20 occurrences; the maximum affix length is four tokens for Chinese/Japanese and two otherwise. Implicit aspects are disabled for English following the benchmark documentation and elsewhere when the training implicit-aspect rate is below 5\%; only Japanese hotel meets the retained policy. Retrieved examples come from the same corpus.

\paragraph{Task 3 fitting details.}
The implemented inventory contains 14 English restaurant categories, 12 categories in the other restaurant corpora, 44 Japanese hotel categories, and 113--121 laptop categories. These counts come from the eligible training pools, not from a canonical label scheme. Fitting samples accumulate complete training reviews until at least 1,000 annotated pairs are covered per corpus. Count features leave out all annotations sharing the sampled review's normalized text. Retrieved demonstrations also omit that text, but the category glossary is built from the whole eligible training pool, so the sampled text is not removed from every glossary entry. Each of the four category variants refits its own fusion weights; ``model + prior'' is therefore not the raw model argmax.

\section{Interpreting the Diagnostics}
\label{app:diagnostics}
The exact-VA diagnostic in Table~\ref{tab:corpora} scores our extracted pairs with gold VA. It removes numerical error on structurally matched predictions and keeps the extracted pairs unchanged, so it bounds what improving VA alone can gain for that pair set; it says nothing about candidate coverage, recall, or category selection. Category accuracy likewise conditions on matched pairs and ignores missing or spurious pairs. In our results, the ratio of Task 3 to Task 2 cF1 is close to this conditional accuracy, but not identical to it, because VA weights and gold tuple counts also enter.

\section{Literature-Audit Protocol}
\label{app:literature-audit}
\input{tables/literature-audit}
Figure~\ref{fig:absa-trend} describes a bounded corpus, not all of ABSA. We searched ACL Anthology metadata for ACL, EMNLP, NAACL, EACL, and COLING main proceedings and associated Findings published in 2004--2025, including LREC-COLING 2024. The case-insensitive title rule requires both ``aspect'' and ``sentiment'', or the standalone abbreviation ``ABSA''. Workshops and demonstrations are outside the scope. The start year follows early feature-level opinion mining \citep{hu2004mining}; matching papers begin in 2008. The Anthology snapshot is commit \texttt{51279f83}, retrieved September 28, 2026.

We additionally searched official AAAI, NeurIPS, and ICML proceedings for 2004--2025 and ICLR conference programs for 2013--2025 with the same rule. AAAI's older directories were retrieved through a public reader proxy; AAAI was not held in 2009. ICLR, NeurIPS, and ICML produced no matching titles, which does not mean they publish no ABSA research. Source URLs, retrieval hashes, track filters, and zero-hit records are preserved in the audit directory.

The combined search returned 293 venue-eligible candidates: 261 from the Anthology and 32 from AAAI. We excluded 16 dataset-only, diagnostic, or non-ABSA-prediction papers, leaving 277 papers. Task or dataset papers remain eligible when they introduce or adapt an actual predictor or training intervention. Each included paper contributes once, regardless of the number of proposed variants or evaluated tasks. Models used only as comparison baselines or mentioned in related work do not affect its category.

\paragraph{Model-use codebook.}
\emph{Generative} means that a proposed method or its tested variant uses a text-generative model at any stage: resource construction, training augmentation, representation extraction, candidate scoring, preprocessing, or task inference. Mixed pipelines count as Generative even when their final task head is discriminative. This includes GPT-, T5-, and BART-family models, translation systems, and autoregressive language-model features such as ELMo and XLNet. Using only the encoder of a text-generative pretrained model also qualifies. The category thus measures model use, not whether a system generates text at inference.

\emph{Discriminative} covers direct label, rating, tag, span, table, or action decisions without identified text-generative model use. BERT/RoBERTa masked-language-model encoders do not qualify as text-generative models under this codebook; neither does masked-token substitution alone. A task-specific pointer or transition decoder without a text-generative language model is not automatically Generative. Early statistical topic models, VAEs, and RBMs also do not qualify merely because they have a probabilistic generative formulation. Latent/discovery approaches are retained as \emph{Other}; a discriminative predictor using a latent auxiliary objective remains Discriminative unless a text-generative model is also used.

We also retain unresolved model use as Other rather than assuming that an undisclosed component is non-generative. For example, \href{https://aclanthology.org/2025.coling-main.269/}{UGTS} names AMRLib and \href{https://aclanthology.org/2021.naacl-main.229/}{GraphMerge} names the Berkeley parser without specifying a checkpoint. Conversely, \href{https://aclanthology.org/2023.acl-long.19/}{APARN} names SPRING, whose documented BART backbone establishes Generative preprocessing. Other contains 11 latent/discovery papers and nine papers with unresolved model use.

Publisher full texts were temporarily unavailable for part of the AAAI expansion. Of its 30 included papers, 12 were checked against full texts, ten against author or associated implementations and dependency documentation, and one topic model against its abstract. Seven remain unresolved; their abstracts establish task eligibility but cannot establish absence of text-generative dependencies. These papers stay in the denominator. Individual records distinguish evidence types and link the inspected sources.

\paragraph{Aggregation and limitations.}
The figure pools papers within five publication periods. Each percentage divides the category count by all included papers in that period, retaining Other in the denominator. Table~\ref{tab:literature-audit} gives annual counts; no observation is imputed for years without matching papers. In 2024--2025, 53 of 82 papers have identified text-generative use and four remain unresolved. Assigning all four to Generative would raise that share from 64.6\% to 69.5\%.

The early 2004--2013 period contains only four papers and cannot establish the field's original method distribution. Unequal period lengths, evolving venue coverage, title vocabulary, and changing task composition further limit interpretation. Coding was model-assisted with targeted source checks, without independent double annotation, so no inter-annotator agreement is available. The accompanying \texttt{analysis/absa-trend/} directory provides titles, links, evidence, exclusions, and the counting protocol. The figure builder computes percentages directly from those records.

\input{sections/prompt-templates}

%% file: tables/dataset.tex
\begin{table*}[t]
\centering\small
\begin{tabular}{@{}lccccccc@{}}
\toprule
Corpus & \begin{tabular}[c]{@{}c@{}}Train\\rev.\end{tabular} & \begin{tabular}[c]{@{}c@{}}T1 dev\\rev.\end{tabular} & \begin{tabular}[c]{@{}c@{}}T1 test\\rev.\end{tabular} & \begin{tabular}[c]{@{}c@{}}T1 test\\ann.\end{tabular} & \begin{tabular}[c]{@{}c@{}}T2 dev\\rev.\end{tabular} & \begin{tabular}[c]{@{}c@{}}T2 test\\rev.\end{tabular} & \begin{tabular}[c]{@{}c@{}}T2 test\\ann.\end{tabular} \\
\midrule
English restaurant & 2284 & 200 & 1000 & 1504 & 200 & 1000 & 2129 \\
English laptop & 4076 & 200 & 1000 & 1421 & 200 & 1000 & 1974 \\
Japanese hotel & 1600 & 200 & 800 & 1092 & 200 & 800 & 1443 \\
Japanese finance & 1024 & 200 & 800 & 1302 & -- & -- & -- \\
Russian restaurant & 1240 & 56 & 1072 & 1637 & 48 & 630 & 1310 \\
Tatar restaurant & 1240 & 56 & 1072 & 1637 & 48 & 630 & 1310 \\
Ukrainian restaurant & 1240 & 56 & 1072 & 1637 & 48 & 630 & 1310 \\
Chinese restaurant & 6050 & 300 & 1000 & 1929 & 300 & 1000 & 2861 \\
Chinese laptop & 3490 & 300 & 1000 & 1673 & 300 & 1000 & 1925 \\
Chinese finance & 1000 & 200 & 842 & 2354 & -- & -- & -- \\
\bottomrule
\end{tabular}
\caption{Dataset statistics (rev.: reviews; ann.: annotations). Tasks 2 and 3 share reviews; Task 3 has one more English laptop test annotation.}
\label{tab:data}
\end{table*}

%% file: tables/literature-audit.tex
\begin{table}[t]
\centering\small
\begin{tabular}{@{}ccccc@{}}
\toprule
Year & D & G & Other & $n$ \\
\midrule
2008 & 0 & 0 & 1 & 1 \\
2010 & 0 & 0 & 1 & 1 \\
2013 & 0 & 0 & 2 & 2 \\
2014 & 1 & 0 & 3 & 4 \\
2015 & 2 & 1 & 1 & 4 \\
2016 & 6 & 1 & 0 & 7 \\
2017 & 2 & 0 & 1 & 3 \\
2018 & 17 & 1 & 2 & 20 \\
2019 & 22 & 0 & 1 & 23 \\
2020 & 29 & 1 & 0 & 30 \\
2021 & 28 & 9 & 4 & 41 \\
2022 & 21 & 10 & 0 & 31 \\
2023 & 9 & 19 & 0 & 28 \\
2024 & 23 & 30 & 1 & 54 \\
2025 & 2 & 23 & 3 & 28 \\
\midrule
Total & 162 & 95 & 20 & 277 \\
\bottomrule
\end{tabular}
\caption{Annual paper counts behind Figure~\ref{fig:absa-trend}. G: the proposed method uses a text-generative model; D: it does not; Other: latent methods or unresolved model use. Years without included papers are omitted.}
\label{tab:literature-audit}
\end{table}

%% file: sections/prompt-templates.tex
\section{Prompt Templates}
\label{app:prompts}
We document the prompt templates used by the final three-task pipeline. Each request consists of a shared \texttt{state} and a dictionary of typed \texttt{questions}; each question specifies its \texttt{type}, \texttt{instructions}, and, for \score{} or \choice{}, \texttt{criteria}. The text below preserves the implemented wording, with line wrapping for presentation. Braced names such as \texttt{\{aspect\}} are substitution slots, not literal input. Corpus-specific reviews, demonstrations, category inventories, and glossaries are filled at runtime.

\newenvironment{prompttext}{\begin{quote}\small\ttfamily\raggedright}{\end{quote}}

\subsection{Shared valence--arousal rubrics}
\label{app:prompt-rubrics}
Both Task 1 and the pair-scoring stage of Task 2 use \texttt{type: score}, with the ordered \texttt{criteria} in Table~\ref{tab:prompt-rubrics}. These nine level descriptions are our rubric; the 1--9 scale and the short dimension definitions below follow the benchmark \citep{lee2026dimabsa}. The returned expected index is zero-based and is shifted by one before calibration (Equation~\ref{eq:score}).

\begin{table*}[t]
\centering\small
\begin{tabularx}{\textwidth}{@{}r>{\raggedright\arraybackslash}X>{\raggedright\arraybackslash}X@{}}
\toprule
Level & Valence criterion & Arousal criterion \\
\midrule
1 & Strongly negative: a severe fault, harsh or contemptuous complaint & Very calm, low energy: the aspect arouses no feeling at all; the writer is indifferent \\
2 & Clearly negative: the aspect is described as bad or disappointing & Calm: the aspect is regarded without emotional charge \\
3 & Moderately negative: real criticism, but not emphatic & Somewhat calm: only the faintest feeling about the aspect \\
4 & Mildly negative: a small complaint or a slight reservation & Mildly calm: a low-energy, subdued feeling \\
5 & Neutral or mixed: no clear polarity, or praise and criticism cancel out & Moderate: an ordinary, middle-of-the-road level of feeling \\
6 & Mildly positive: a small or lukewarm compliment & Moderately activated: the feeling runs a little above ordinary \\
7 & Moderately positive: the aspect is described as good & Activated, excited: a clearly energised feeling about the aspect \\
8 & Clearly positive: strong approval, the aspect is praised & Strongly activated: high energy, intensely felt \\
9 & Strongly positive: enthusiastic praise, superlatives, delight & Extremely activated, high energy: furious or thrilled; the strongest feeling \\
\bottomrule
\end{tabularx}
\caption{Verbatim ordered criteria for the shared \score{} questions. Level numbers show the benchmark scale; the API criterion indices are 0--8.}
\label{tab:prompt-rubrics}
\end{table*}

\paragraph{Dimension definitions.}
The corresponding sentence is appended to the question:
\begin{prompttext}
Valence: 1 = most negative, 9 = most positive.
\end{prompttext}
\begin{prompttext}
Arousal: 1 = calm/low intensity, 9 = excited/high intensity.
\end{prompttext}
\paragraph{Dimension-specific focus.}
The \texttt{instructions.focus} field ends with the following text for valence and arousal, respectively:
\begin{prompttext}
Judge only the sentiment directed at this aspect; ignore sentiment toward any other aspect in the text.
\end{prompttext}
\begin{prompttext}
Judge only the feeling directed at this aspect; ignore feeling toward any other aspect in the text. Arousal is how activated that feeling is -- how calm or how excited -- not how positive or negative it is.
\end{prompttext}
\paragraph{Implicit aspects.}
For VA questions, \texttt{the aspect "\{aspect\}"} becomes the following phrase when the aspect is \texttt{NULL}:
\begin{prompttext}
the aspect that is left implicit and never named in the text
\end{prompttext}
\subsection{Task 1: given-aspect regression}
\label{app:prompt-task1}
The \texttt{state} has two fields: \texttt{review\_to\_score} contains the input review, and \texttt{labelled\_examples} contains nine fixed training demonstrations. Each demonstration has \texttt{review}, \texttt{aspect}, \texttt{valence}, and \texttt{arousal} fields; the two numeric labels are rounded to two decimals. Selection follows Section~\ref{sec:method}.

\paragraph{Questions.}
For each given aspect, the two \texttt{instructions.question} fields begin as follows:
\begin{prompttext}
What valence given the aspect "\{aspect\}" in `review\_to\_score`?
\end{prompttext}
\begin{prompttext}
What arousal given the aspect "\{aspect\}" in `review\_to\_score`?
\end{prompttext}
Each question then appends its dimension definition from Appendix~\ref{app:prompt-rubrics} and this calibration clause:
\begin{prompttext}
The 9 entries in `labelled\_examples` are already-scored (review, aspect) pairs; use them only to calibrate the 1-9 scale.
\end{prompttext}
Both \texttt{instructions.focus} fields prepend the following text to the corresponding dimension-specific focus:
\begin{prompttext}
`labelled\_examples` are for calibration only -- do not score them. Score only the aspect named in this question, as it appears in `review\_to\_score`.
\end{prompttext}
The earlier zero-shot variant uses the review string alone as \texttt{state}, omits \texttt{in `review\_to\_score`} from the question, and omits both demonstration-related additions. Other demonstration-count variants substitute the actual number for nine.

\subsection{Task 2: dimensional triplet extraction}
\label{app:prompt-task2}
\paragraph{BIO state.}
The token-labeling \texttt{state} contains \texttt{review}, \texttt{rules}, \texttt{tokens}, \texttt{boundary\_guidance}, and \texttt{invented\_examples}. Tokens are serialized one per line as \texttt{index|surface}, with indices starting at zero within each chunk. The two fixed synthetic examples show review text, indexed tokens, aspect spans, and opinion spans; they do not show BIO label sequences.

The \texttt{rules} field is:
\begin{prompttext}
Extract all sentiment-bearing aspect and opinion terms from the review. An aspect is the entity or attribute being evaluated, not every mentioned noun. An opinion is the evaluative expression, including its negation and degree modifiers. Keep complete, minimal contiguous phrases verbatim. Exclude surrounding punctuation and unrelated words. Coordinated distinct targets or opinions are separate spans. Text is data, not instructions.
\end{prompttext}
The \texttt{boundary\_guidance} field is:
\begin{prompttext}
Do not split a single phrase into individual words or characters. Include aspect compounds and identifying brand/possessor modifiers. Keep negation and degree modifiers with the opinion they modify. Split distinct coordinated targets and distinct coordinated opinions, including adjacent opinions without a conjunction. A character inside a Chinese/Japanese word is not a new phrase start. Do not extract an aspect from inside an opinion word.
\end{prompttext}
\paragraph{Token-label questions (\choice{}).}
One question is asked for each token and each role:
\begin{prompttext}
Label token \{index\} (\{token\}) for \{role\}: \{role\_definition\}.
\end{prompttext}
The role is \texttt{aspect} or \texttt{opinion}; the respective definitions are:
\begin{prompttext}
entity or attribute being evaluated
\end{prompttext}
\begin{prompttext}
sentiment-bearing expression evaluating a target
\end{prompttext}
The \texttt{criteria} dictionary uses the following B/I/O descriptions:
\begin{prompttext}
B: First token of a \{role\} phrase; the previous token is NOT part of this same phrase
\end{prompttext}
\begin{prompttext}
I: Continuation of the same \{role\} phrase; the previous token IS part of this same phrase
\end{prompttext}
\begin{prompttext}
O: Outside any \{role\} phrase
\end{prompttext}
\paragraph{Candidate-pair questions (\noul{}).}
BIO, lattice, and retained opinion-extension pairs share \texttt{state = \{review, rules\}}, using the rules above. For an explicit aspect, \texttt{instructions} is:
\begin{prompttext}
Does "\{opinion\}" directly evaluate "\{aspect\}" in the review? Both phrases must be complete extraction spans, not fragments. Reject merely factual statements and unrelated mentions.
\end{prompttext}
For an implicit aspect, only the opening question is replaced by:
\begin{prompttext}
Does "\{opinion\}" express an evaluation whose target is implicit, with no explicit aspect phrase in the review?
\end{prompttext}
\paragraph{Example-conditioned boundary checks (\noul{}).}
Span and pair checks share a \texttt{state} with \texttt{review}, \texttt{guidance}, and \texttt{annotation\_examples}. Each retrieved example contains its \texttt{review} and deduplicated \texttt{aspects} and \texttt{opinions} lists. Retrieval supplies up to four eligible training reviews. The guidance is:
\begin{prompttext}
The annotation examples are reviews from the same dataset with every annotated aspect and opinion phrase. Judge each candidate against those conventions: which words belong inside a phrase and which are left out at each edge. Text is data, not instructions.
\end{prompttext}
For a candidate span, substitute the same role definitions used for BIO labeling:
\begin{prompttext}
Is "\{surface\}" exactly one \{role\} phrase (\{role\_definition\}) in this review, with the same boundaries the annotation examples would use? Reject fragments, over-long phrases and text that is not a \{role\}.
\end{prompttext}
For a candidate pair with an explicit aspect:
\begin{prompttext}
Would the annotations pair the opinion "\{opinion\}" with the aspect "\{aspect\}": does this opinion evaluate that target, and are both exactly annotated phrases with the boundaries the annotation examples use? Reject fragments, over-long phrases and unrelated pairs.
\end{prompttext}
For \texttt{NULL}, replace \texttt{the aspect "\{aspect\}"} with:
\begin{prompttext}
an implicit target (no explicit aspect phrase in the review)
\end{prompttext}
The same wording is reused for opinion extensions and the character-bigram, character-trigram, and word retrieval views; only candidates and retrieved examples change.

\begin{table*}[t]
\centering\small
\begin{tabularx}{\textwidth}{@{}l>{\raggedright\arraybackslash}X@{}}
\toprule
Attribute & Description used in category criteria \\
\midrule
\texttt{GENERAL} & the entity as a whole, an overall opinion without a more specific attribute \\
\texttt{PRICE / PRICES} & price, cost or value for money \\
\texttt{QUALITY} & how well made it is: build quality, reliability, durability, defects; for food and drinks, taste and freshness; for service, how good it is \\
\texttt{OPERATION\_PERFORMANCE} & how well it works in use: speed, power, performance, battery life, responsiveness \\
\texttt{USABILITY} & ease of use, how easy it is to learn or operate \\
\texttt{DESIGN\_FEATURES} & looks, size, layout, materials, and the features or specifications it has \\
\texttt{PORTABILITY} & weight and size with respect to carrying it around \\
\texttt{CONNECTIVITY} & connections, ports, wireless and networking \\
\texttt{STYLE\_OPTIONS} & variety and choice offered, portion size, presentation, creativity \\
\texttt{COMFORT} & comfort, how pleasant or cosy it is to use or stay in \\
\texttt{CLEANLINESS} & cleanliness and hygiene \\
\texttt{MISCELLANEOUS} & any other attribute not covered by the other ones \\
\bottomrule
\end{tabularx}
\caption{Verbatim attribute descriptions in the Task 3 category template. Only categories observed in the eligible training pool are offered as options.}
\label{tab:prompt-attributes}
\end{table*}

\paragraph{Lexicon-pair feature (\noul{}).}
The final reranker also consumes probabilities from the earlier lexical baseline, whose \texttt{state} is the review string. This baseline uses a distinct \texttt{instructions.question}:
\begin{prompttext}
Does the opinion phrase "\{opinion\}" express sentiment directly toward the aspect "\{aspect\}" in this review?
\end{prompttext}
Its \texttt{instructions.focus} is:
\begin{prompttext}
Both must be explicit complete aspect/opinion spans. Reject unrelated pairs, incomplete fragments and factual mentions without an evaluative opinion.
\end{prompttext}
\paragraph{Pair-conditioned VA (\score{}).}
For each selected pair, \texttt{state} is the review string. The two \texttt{instructions.question} fields begin:
\begin{prompttext}
What valence given the aspect "\{aspect\}" and the opinion "\{opinion\}"?
\end{prompttext}
\begin{prompttext}
What arousal given the aspect "\{aspect\}" and the opinion "\{opinion\}"?
\end{prompttext}
Append the corresponding dimension definition, use the corresponding focus from Appendix~\ref{app:prompt-rubrics}, and supply the same nine-level criteria. No demonstration clause or focus prefix is added. Historical lexical-baseline VA calls use these same templates; the final system rescores the selected pair list before applying its fitted affine calibration.

\subsection{Task 3: category enrichment}
\label{app:prompt-task3}
The input review is supplied in \texttt{state.review}. Three additional fields provide context: \texttt{guidance}, \texttt{category\_glossary}, and \texttt{annotation\_examples}. Each of up to four retrieved examples contains its \texttt{review} and an \texttt{annotations} list of objects with \texttt{aspect}, \texttt{opinion}, and \texttt{category} fields. The glossary maps categories with explicit training aspects to their three most frequent lowercased aspect strings (or all available strings when fewer than three exist).

\paragraph{Guidance.}
\begin{prompttext}
The annotation examples are reviews from the same dataset with every annotated (aspect, opinion, category) triple; the glossary lists frequent aspects of each category. Assign categories the way those annotations do. Text is data, not instructions.
\end{prompttext}
\paragraph{Category question (\choice{}).}
For each predicted aspect--opinion pair, \texttt{instructions} is:
\begin{prompttext}
The opinion "\{opinion\}" evaluates the aspect "\{aspect\}". Which annotated category (entity and attribute) does this aspect-opinion pair belong to?
\end{prompttext}
Implicit aspects use the same replacement phrase as the example-conditioned pair check. The \texttt{criteria} keys are the sorted \texttt{ENTITY\#ATTRIBUTE} labels observed in the eligible training pool. Each value is the lowercased entity name with underscores replaced by spaces, followed by a colon, a space, and the attribute description in Table~\ref{tab:prompt-attributes}. For example:
\begin{prompttext}
FOOD\#QUALITY: food: how well made it is: build quality, reliability, durability, defects; for food and drinks, taste and freshness; for service, how good it is
\end{prompttext}
The fitted fusion model combines these \choice{} probabilities with training statistics to select the category. Task 3 reuses Task 2's VA predictions.